\documentclass[10pt,twocolumn,letterpaper]{article}

\usepackage[pagenumbers]{cvpr} % To force page numbers, e.g. for an arXiv version

\usepackage{graphicx}
\usepackage{amsmath}
\usepackage{amssymb}
\usepackage{booktabs}
\usepackage{censor}
\usepackage{enumitem}
\usepackage{subcaption}
\usepackage[accsupp]{axessibility}
\usepackage[pagebackref,breaklinks,colorlinks]{hyperref}
\usepackage{nameref}
\usepackage{xcolor}

\usepackage[capitalize]{cleveref}
\crefname{section}{Sec.}{Secs.}
\Crefname{section}{Section}{Sections}
\Crefname{table}{Table}{Tables}
\crefname{table}{Tab.}{Tabs.}

\def\cvprPaperID{10} % *** Enter the CVPR Paper ID here
\def\confName{CVPR}
\def\confYear{2022}
\newcommand{\dname}[0] {Visuelle 2.0\xspace}
\newcommand{\dnameo}[0] {Visuelle}
\newcommand{\dcomp}[0] {Nuna Lie}

\DeclareRobustCommand{\myparagraph}[1]
{\noindent\textbf{#1}} 

\begin{document}

%%%%%%%%% CONCETTI FIGHI
% inventory backlog
% zara super responsive
\title{The multi-modal universe of fast-fashion: the \dname{} benchmark}

\author{
Geri Skenderi\textsuperscript{1}%$^*$
\quad
Christian Joppi\textsuperscript{2}%$^*$
\quad
Matteo Denitto\textsuperscript{2}%$^*$
\quad
Berniero Scarpa\textsuperscript{3}
\quad
Marco Cristani\textsuperscript{1,2}
\vspace{1ex}
\\
\textsuperscript{1} Universit\`a degli Studi di Verona \quad
\textsuperscript{2} Humatics s.r.l. 
\quad
\textsuperscript{3} Nuna Lie s.r.l. 
}
\maketitle
%\def\thefootnote{*}\footnotetext{Indicates equal contribution}
%\def\thefootnote{\arabic{footnote}}

%%%%%%%%% ABSTRACT
\begin{abstract}
    We present \dname{}, the first dataset useful for facing diverse prediction problems that a fast-fashion company has to manage routinely. Furthermore, we demonstrate how the use of computer vision is substantial in this scenario. \dname{} contains data for 6 seasons / 5355 clothing products of \emph{\dcomp}\footnote{\url{http://www.nunalie.it}}, a famous Italian company with hundreds of shops located in different areas within the country. In particular, we focus on a specific prediction problem, namely \emph{short-observation new product sale forecasting} (SO-fore). SO-fore assumes that the season has started and a set of new products is on the shelves of the different stores. The goal is to forecast the sales for a particular horizon, given a short, available past (few weeks), since no earlier statistics are available. To be successful, SO-fore approaches should capture this short past and exploit other modalities or exogenous data. To these aims, \dname is equipped with disaggregated data at the item-shop level and multi-modal information for each clothing item, allowing computer vision approaches to come into play. The main message that we deliver is that the use of image data with deep networks boosts performances obtained when using the time series in long-term forecasting scenarios, ameliorating the WAPE and MAE by up to 5.48\% and 7\% respectively compared to competitive baseline methods. The dataset is available at : \url{https://humaticslab.github.io/forecasting/visuelle}.
\end{abstract}

\section{Introduction}
\label{sec:intro}
\begin{figure}
        \includegraphics[width=\linewidth]{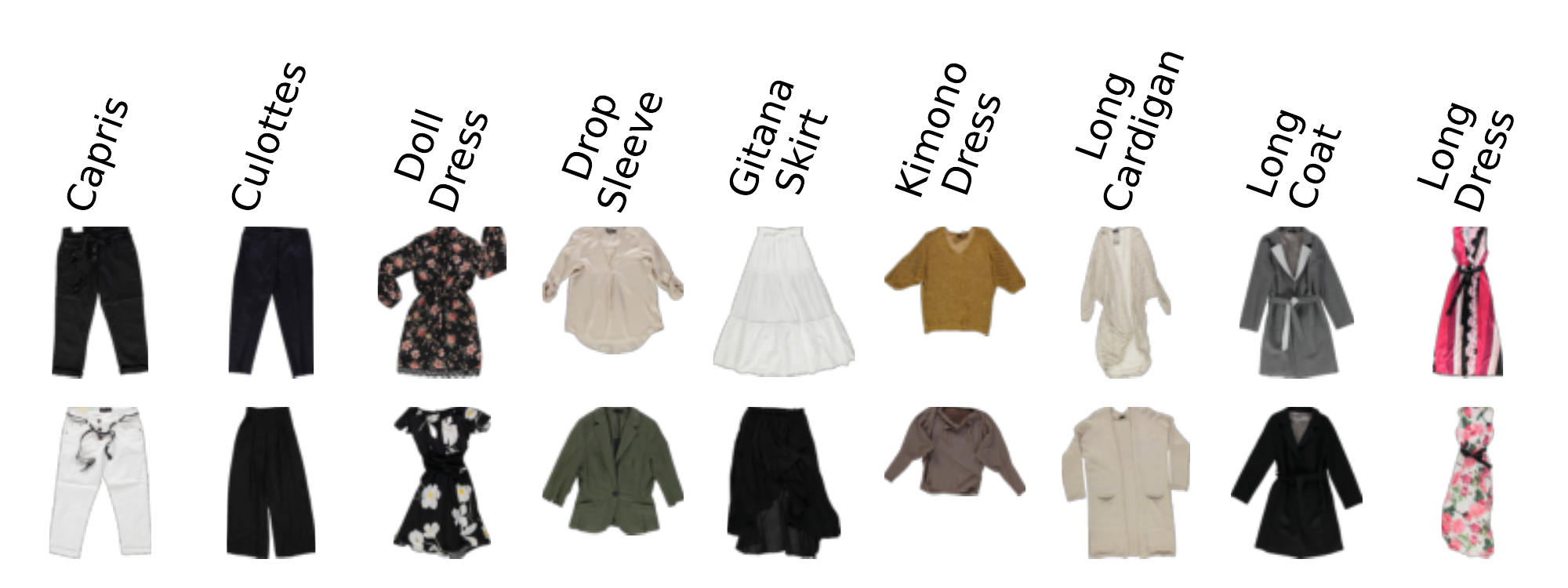}
        \includegraphics[width=\linewidth]{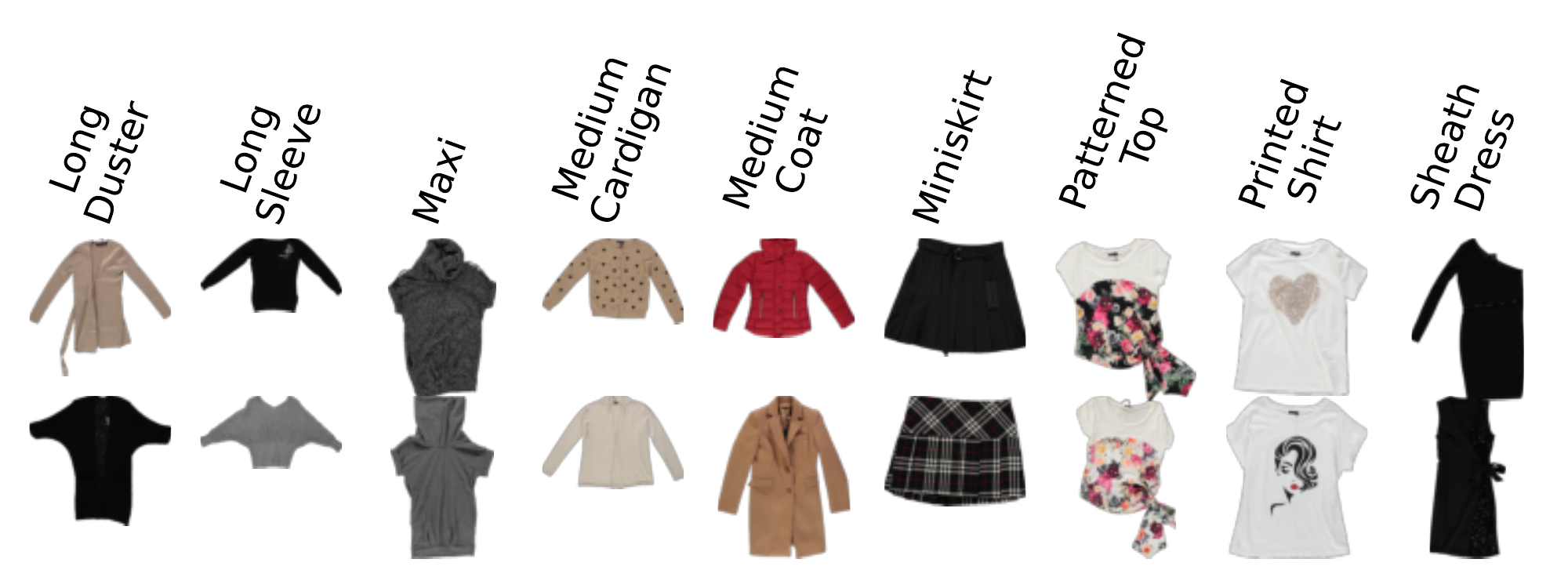}
        \includegraphics[width=\linewidth]{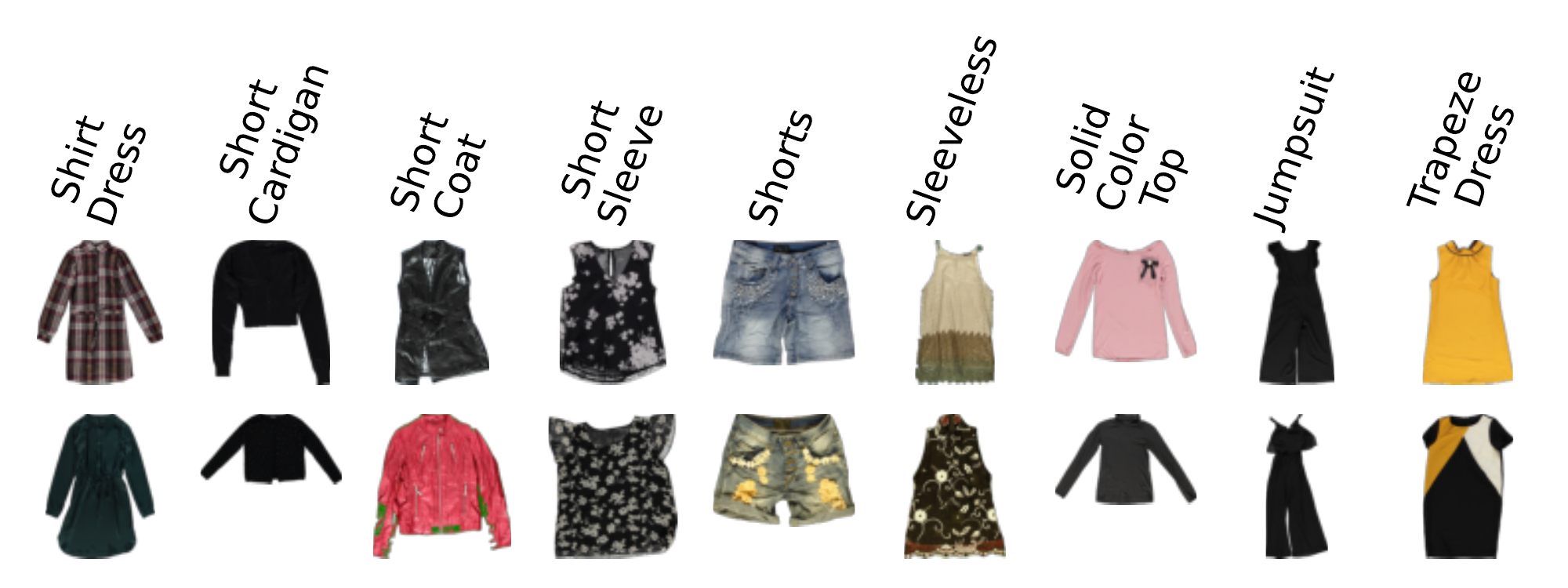}
        \caption{\textbf{Why are images important for fashion forecasting?}. A visual excerpt of \dname{}, organized per category, which shows the crucial role of images: in the \emph{patterned shirt category} two products have exactly the same textual attributes (floreal), so the image becomes necessary for discriminative tasks. Similar considerations hold for \emph{short sleeves} and \emph{doll dress}.}
        \label{fig:mosaic}
        \vspace{-0.5cm}
\end{figure}

Fashion forecasting has traditionally been studied in scientific sectors other than computer vision, such as operational research and logistics, with the primary aim of predicting trends~\cite{holland2017fashion,kim2021fashion}, sales~\cite{loureiro2018exploring,ma2021retail,kharfan2021data}, and performing demand forecasting~\cite{sajja2021explainable,feizabadi2022machine}. In recent years this trend has faded, showing an increasing cross-fertilization with computer vision ~\cite{chang2021fashion,al2017fashion,ma2020knowledge,Mall2019GeoStyle,ding2021leveraging}. 

In this paper, we present \dname, which contains real data for 5355 clothing
products of a retail fast-fashion Italian company, \emph{\dcomp}.
For the first time ever, a retail fast-fashion company has decided to share part of its data to provide a genuine benchmark for research and innovation purposes. Specifically, \dname{} provides data from 6 fashion seasons (partitioned in Autumn-Winter and Spring-Summer) from 2016-2020, right before the Covid-19 pandemic\footnote{The pandemic represented an unicum in the dynamics of the fast fashion companies, so it has not been included. The market effectively restarted in AW 21-22 which is an ongoing season at the time of writing.}. Each product in our dataset is accompanied by an HD image, textual tags and more. The time series data are disaggregated at the shop level, and include the sales, inventory stock, max-normalized prices\footnote{Prices have been normalized for the sake of confidentiality.} and discounts. This permits to perform SO-fore considering each single store. Exogenous time series data is also provided, in the form of Google Trends based on the textual tags and multivariate weather conditions of the stores' locations. Finally, we also provide purchase data for 667K customers whose identity has been anonymized, to capture personal preferences. With these data, \dname{} allows to cope with several problems which characterize the activity of a fast fashion company: \emph{new product demand forecasting}~\cite{sajja2021explainable,feizabadi2022machine}, \emph{short-observation new product sales forecasting}~\cite{loureiro2018exploring,ma2021retail,kharfan2021data} and \emph{product recommendation}~\cite{chakraborty2021fashion}. 

\begin{figure}[t]
	\centering
        \includegraphics[width=0.9\linewidth]{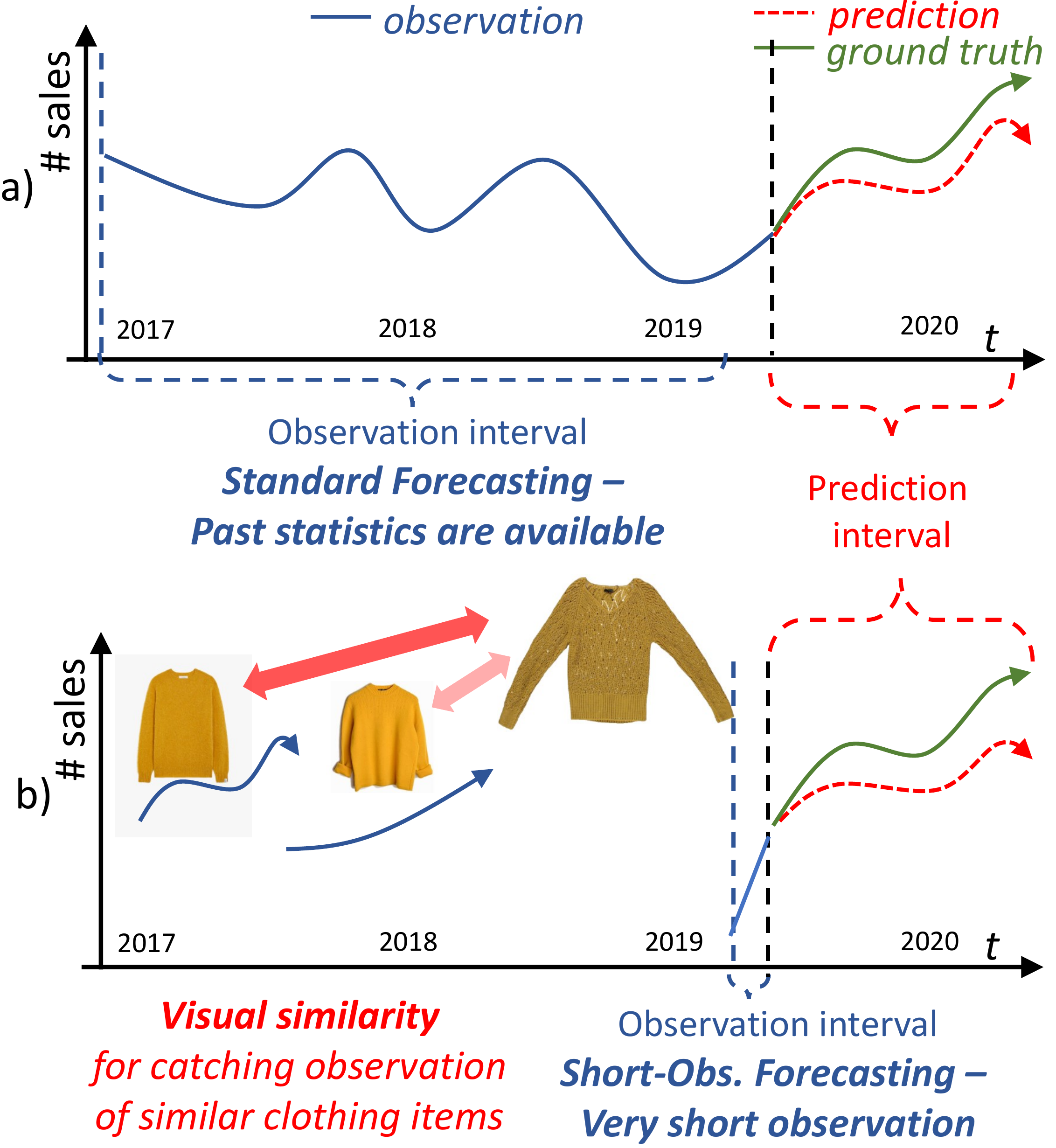}
        \caption{Short-observation new product sales forecasting (SO-fore): in a), the standard forecasting setup is reported. In b), SO-fore is sketched, focusing on a very short observation window (2 weeks) to predict sales. Here we show that relying on visual similarity to extract statistics of similar data improves forecasting.}
        \vspace{-0.2cm}
        \label{fig:teaser}
        %\vspace{-0.5cm}
\end{figure}
In this paper, we focus on one of these problems: short-observation new product forecasting (SO-fore). SO-fore aims at predicting the future sales in the short term, having a past statistic given by the early sales of a given product (Fig.~\ref{fig:teaser}b provides a visual comparison with standard forecasting in Fig.~\ref{fig:teaser}a). In practice, after a few weeks from the delivery on market, one has to sense how well a clothing item has performed and forecast its behavior in the coming weeks. This is crucial to improve \emph{restocking policies}~\cite{maass2014improving}: a clothing item with a rapid selling rate should be restocked to avoid stockouts. Two particular cases of the SO-fore problem will be taken into account: SO-fore$_{2-10}$, in which the observed window is 2 weeks long and the forecasting horizon is 10 weeks long, required when a company wants to implement few restocks~\cite{fisher2001optimizing}; SO-fore$_{2-1}$, where the forecasting horizon changes to a single week, and is instead required when a company wants to take decisions on a weekly basis, as in the \emph{ultra-fast fashion} supply chain~\cite{taplin2014global,choi2014fast}. 

Our findings show that the usage of image data is crucial in absence of long term statistics which characterize the sales, because the pictorial content of a fast fashion product can be used to inherit long term statistics \emph{from visual similarity} (see Fig.~\ref{fig:teaser}b): the images allow to refer with high precision to past data that are akin to the product of analysis, providing informative priors.

\dname is a substantial extension of the unpublished \dnameo{} dataset~\cite{skenderi2021well}, used only for \emph{new product demand forecasting}, where less data (no weather conditions, no customer data) was furnished, aggregated per-product over all the retail stores (no geographical/store dimension).

\section{The Dataset} \label{sec:dataset}
\dname{} describes the sales between Nov. 2016 and Dec. 2019 of 5355 different products across 110 different shops. For each product, multi-modal information is available, as described in the following.

\begin{figure}
    \centering
    \subfloat[]{\includegraphics[width=0.49\linewidth]{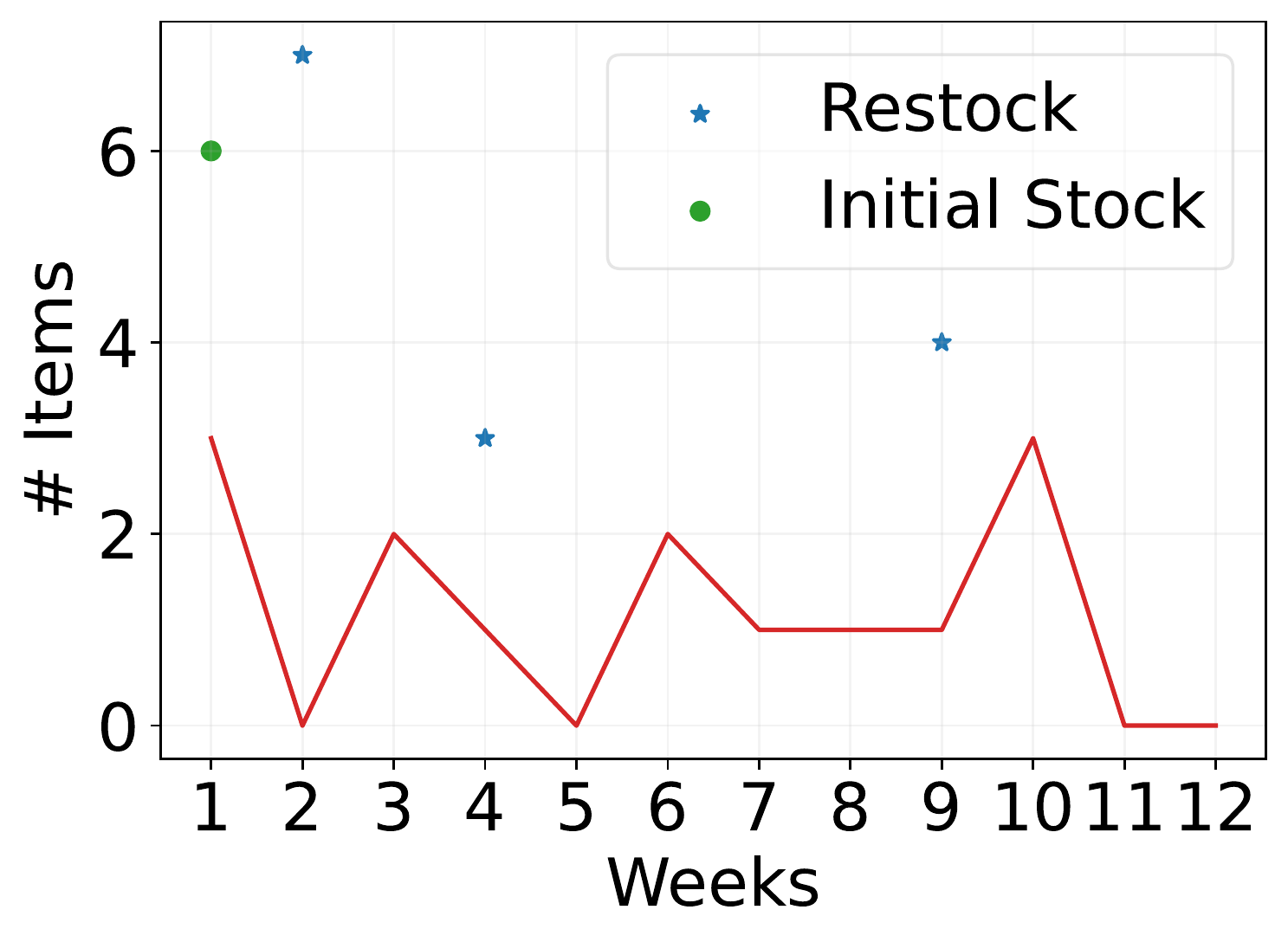}}
    \subfloat[]{\includegraphics[width=0.49\linewidth]{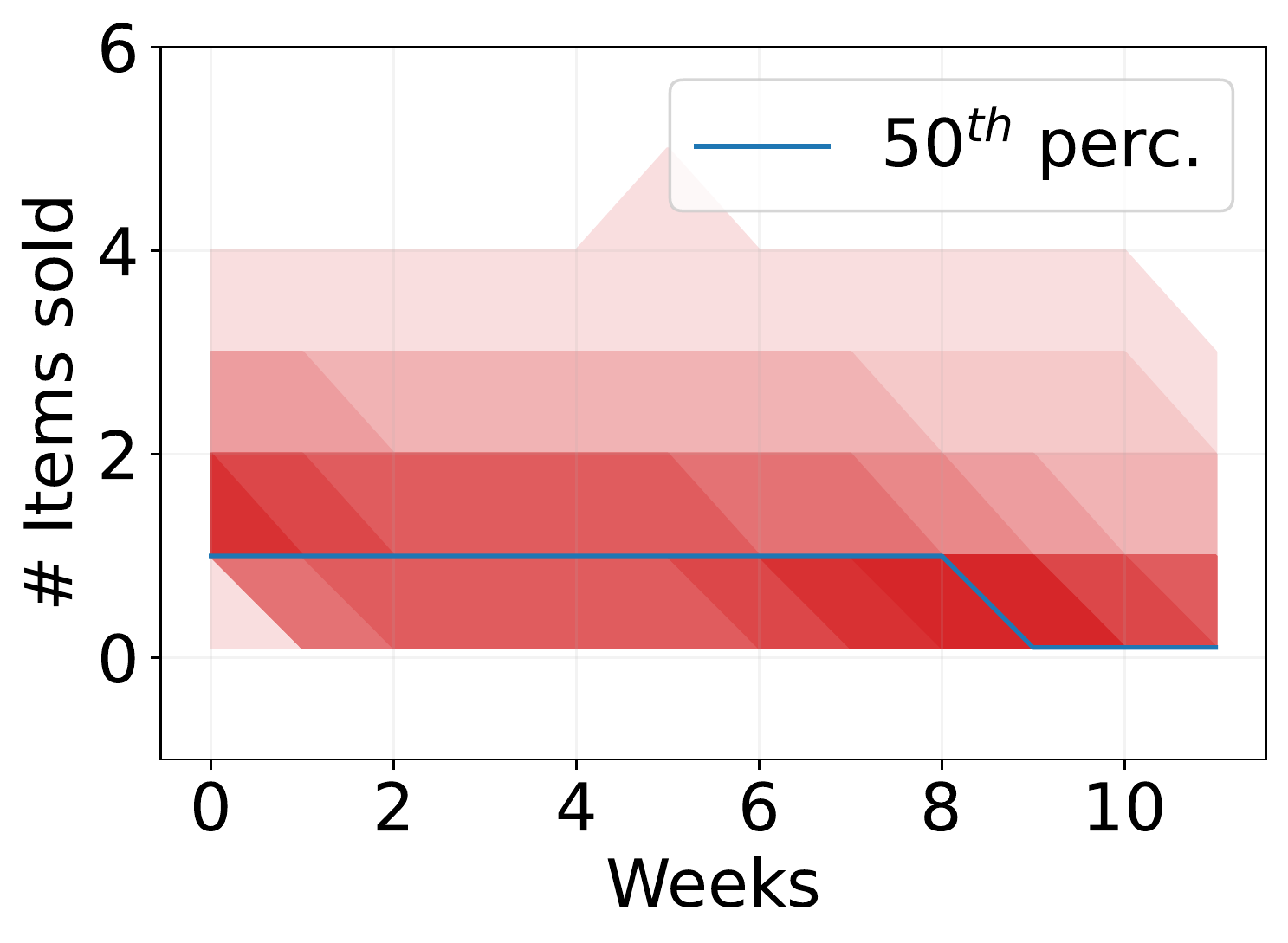}}
    \vspace{-0.2cm}
  \caption{Time series in \dname{}; a) an example of a sale signal, with the restocks highlighted by the asterisks; b) the [0;$95^{\text{th}}$] percentile statistics of the sales signals related to the season SS19, considering all the stores and all the products. }
    \label{fig:sale_stat}
    \vspace{-0.5cm}
\end{figure}

\myparagraph{Time series data.}
Given a product $i$ of size $s$  at a retail store $r$, we refer to its \emph{product sale} signal as $S(i,s,r,t)$ where $t$ refers to the $t$-th week of market delivery, with $i=1,...,N$, $s=1,...,M$, $r=1,...,L$ and $t=1,...,K$. We also define the \emph{inventory position} signal $I(i,s,r,t)$ indicating the inventory on-hand %+ orders outstanding - b
for that quadruplet $(i,s,r,t)$. Combining these data, we can individuate all those \emph{legit} sales signals that do not involve a stockout until the $K$-th week. Formally, a legit sale signal is $S(i,s,r,1),...,S(i,s,r,t_{legit})$ where $t_{legit}+1$ indicates the first week with a stockout $I(i,s,r,t_{safe}+1) = 0$, and $t_{safe}>K$. Hence, we guarantee that a zero-sale legit signal (i.e. $S(i,s,r,t)=0$) is provided \emph{only} when nobody bought $i$ (even tough it was available at the shop) and not because of a stockout. These signals are important because they focus on the net performance of a product, independently on the inventory management. To make the signals denser, we aggregate the different sizes obtaining the final $sale(i,r,t)=\sum_{s=1}^{M}A(i,s,r,t)$.
Additionally, we include the \emph{Restock flag} signal $R(i,s,r,t)$ indicating when a restocking has been carried out. Fig.~\ref{fig:sale_stat}a reports one example of a legit sale signal. The initial stock is represented by the initial amount of items available in the inventory.  Fig.~\ref{fig:sale_stat}b depicts a log-density plot of the legit sales of all the products, averaged over categories and retail stores during the SS19 season. Fig.~\ref{fig:retail_stat} shows sales statistics for the 110 available shops.

\begin{figure}
        \centering
        \includegraphics[width=0.9\linewidth]{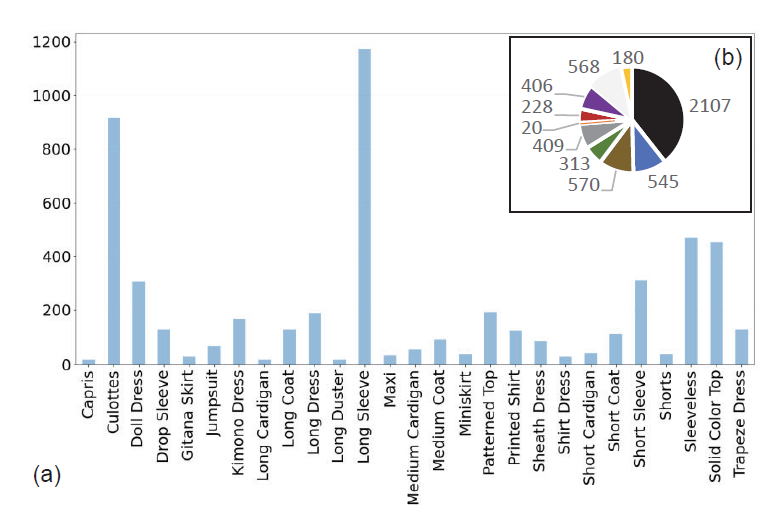}
        \vspace{-0.3cm}
        \caption{Cardinality of products in \dname{} by (a) categories, (b) colors}
        \label{fig:cardinalities}
        \vspace{-0.3cm}
\end{figure}

\myparagraph{Image data} 
Each product is associated with an RGB image that has a resolution which varies from 256x256 to 1193x1172, with a median size of 575x722 (WxH). %Images have been captured in a controlled environment, in order to avoid color inaccuracies and potential biases in the predictions~\cite{nitse2004impact}.
Each image portrays the clothing item on a white background, with no person wearing it. Some examples of these images are provided in Fig.~\ref{fig:mosaic}.

\begin{figure}
\centering
        \includegraphics[width=0.9\linewidth]{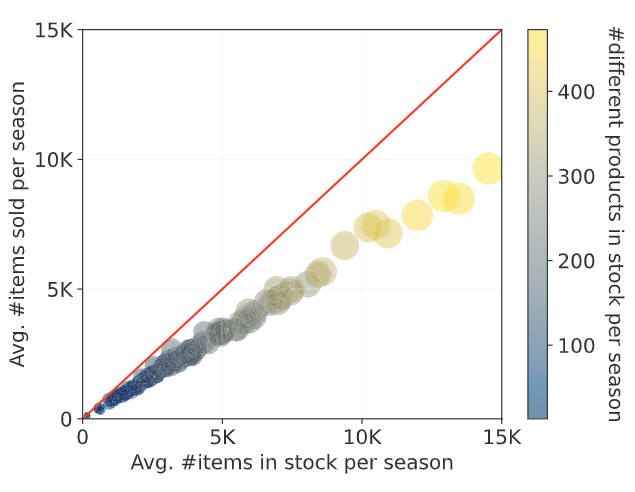}
        \caption{Retail stores sales statistics; each blob indicates the number of items received in their inventory (x-axis) VS number of items sold. The blob size indicates the number of \emph{different} products in the inventory. The red line represents an ideal performance (sales=inventory). The figure says small shops tend to have better performances.}
        \vspace{-0.2cm}
        \label{fig:retail_stat}
        \vspace{-0.4cm}
\end{figure}

\begin{figure}
        \subfloat[]{\includegraphics[width=0.49\linewidth]{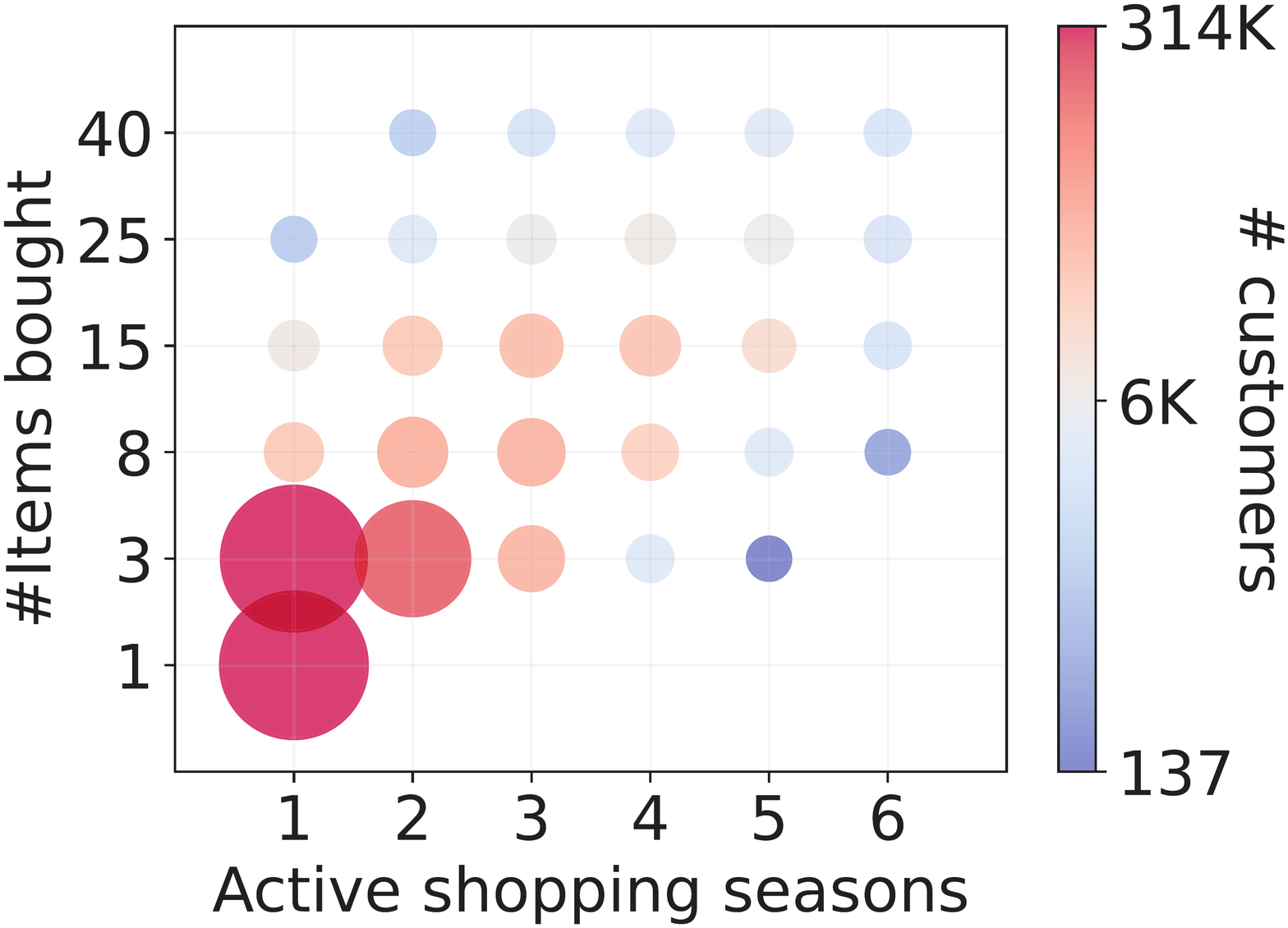}}
        \subfloat[]{\includegraphics[width=0.49\linewidth]{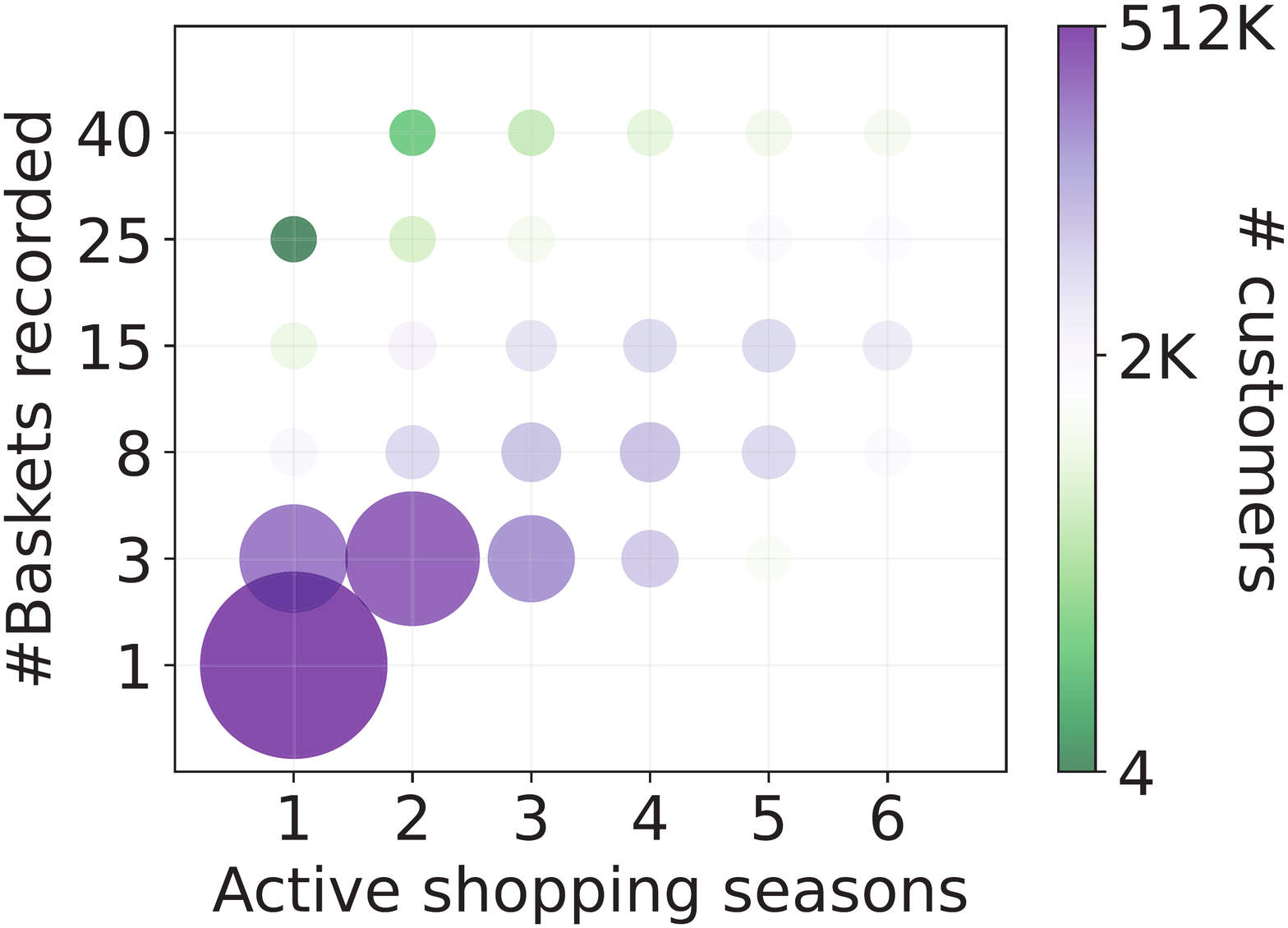}}
        \vspace{-0.2cm}
        \caption{\textbf{Customer shopping history statistics}: a) the blob size indicates how many customers have a shopping history of $n$ seasons long (even non consecutive seasons, at least one item bought per ``active'' season) on the x-axis, consisting of $m$ items in total on the y-axis; b) Number of baskets associated to customers over shopping seasons.}
        \label{fig:customer_hindex}
        \vspace{-0.5cm}
\end{figure}

\myparagraph{Text data} 
Multiple textual tags related to each product's visual attributes are available. These tags have been extracted with diverse procedures or chosen by hand, carefully validated by the Nuna Lie team. The first tag is the \emph{category}, taken from a vocabulary of 27 elements, visualized in Fig.~\ref{fig:cardinalities}a; the cardinality of the products shows large variability among categories overall, due to the fact that some categories (e.g. long sleeves) cost less and ensure higher earnings. The \emph{color} tag (Fig.~\ref{fig:cardinalities}b) represents the most dominant color chosen among 10 manually detected colors. The \emph{fabric} tag comes directly from the technical sheet of the products, chosen from a vocabulary of 59 elements. Finally, the release date for each product-shop pair is recorded as a textual string.\\
\myparagraph{Customer purchase data} 
\dname{} contains anonymized data for 667086 customers, who have requested a fidelity card thanks to which it is possible to extract the history of their purchases and the baskets of products they bought. These data consists of: ID of the purchased product, date-time of purchase, retail store ID and quantity. Fig.~\ref{fig:customer_hindex} gives a glimpse on the distribution of these data, where it is possible to note that there are around 6k  users which have bought continuously a total of 25 products over 4 seasons (Fig.~\ref{fig:customer_hindex}a). More than 2K users have bought continuously 15 baskets of products over 4 seasons. Customer data are useful to test recommendation approaches, whose goal is to recommend available products that the user will eventually buy, possibly exploiting image data to capture personal aesthetic preferences. In this paper, we do not tackle this problem.   
\\\myparagraph{Exogenous time series}
\emph{Google Trends} time series for each product are provided, based on the product's three associated attributes: ${color,category, fabric}$. The trends are downloaded starting from 52 weeks before the product's release date, essentially providing a popularity curve for each of the attributes. Google Trends' efficacy for predictions on fast fashion problems has been demonstrated before in~\cite{silva2019googling,skenderi2021well}. \emph{Weather reports} downloaded from IlMeteo\footnote{\url{https://www.ilmeteo.it/portale/archivio-meteo/}} are also supplied, containing the real weather conditions on a daily basis at the municipality level. The efficacy of weather reports for forecasting in fast fashion has been demonstrated in~\cite{beheshti2015survey,sull2008fast}.

\section{Experiments} \label{sec:exp}
Here we show how \dname{} is a genuine benchmark for two types of SO-fore: 
i) SO-fore$_{2-10}$ and ii) SO-fore$_{2-1}$. 

\textbf{SO-fore$_{2-10}$} allows the company to customize the restocking operations for each product on the basis of the early sales, minimizing the number of such operations. Firstly, the sales series are split into an observation window and a horizon window (i.e. the known past and the future to forecast), set here to 2 and 10 weeks respectively, covering the 12-week fast fashion life-cycle~\cite{thomassey2016intelligent}. Formally, the goal is to perform a multi-step forecast of the sales of an item $i$ in a store $r$ ($sale(i,r,3), ..., sale(i,r,12)$), given the first two time-steps. Two weeks is a standard period to sufficiently understand the whereabouts of the fashion market and take decisions for the future~\cite{yu2008web,thomassey2014sales}. 

\textbf{SO-fore$_{2-1}$} serves in other contexts where a weekly restocking schedule is adopted~\cite{taplin2014global,choi2014fast}.
The idea is to estimate the time-point $sale(i,r,t)$ given the previous two time-steps $sale(i,r,t-1)$, $sale(i,r,t-2)$. Similarly to before, we set the initial observation window to 2, but make a unique prediction corresponding to each observed window.

In both cases we split the data into train and test sets, where the test set contains the 10\% most recent item-shop pairs, such that the items that are seen in training will have always been released before the ones we test on. We use the following three approaches as baselines: 
\begin{itemize}[noitemsep, nolistsep, leftmargin=*]
    \item Classical forecasting methods, namely the Naive method~\cite{FPAP2} (using the last observed ground truth value as the forecast) and Simple Exponential Smoothing (SES)~\cite{hyndman2008smoothing,FPAP2};
    \item kNN~\cite{Martinez2019knn,ekambaram2020attention}, which produces forecasts based on product similarity. This is done by finding k-Nearest Neighbors from the past that are similar to the input product and performing a weighted average of their sales. We set $k=11$ and we compute the similarity between products using the known time series or image features extracted by a pre-trained ResNet~\cite{He_2016_CVPR};
    \item An autoregressive, attention-based RNN architecture \cite{ekambaram2020attention}, where the different data modalities are first processed separately and then merged together through several additive attention modules \cite{Bahdanau2015}. 
\end{itemize}
Results are displayed in Table~\ref{tab:st-res39} and Table~\ref{tab:st-res31}, where all the listed approaches only use sales time series data as input, unless specified otherwise. Classical forecasting approaches tend to give poor performances due to the small number of observations~\cite{FPAP2}. The kNN-based methods show an improvement over the statistical forecasting baselines, demonstrating that inter-product similarity is important when predicting future sales. This is tied with the notion that new products will sell comparably to older, similar products. Trivially utilizing the images with kNN lowers the performances, but their contribution is highlighted when utilising a neural network architecture. Cross-Attention RNN\cite{ekambaram2020attention} outperforms the others by a noticeable margin, because the model is able to learn non-linear, inter-product dependencies throughout the whole training set and also advanced temporal dynamics. It is worth noting that we reach the best performances in SO-fore$_{2-10}$ by pairing each product's time series input with its respective image. This shows that visual representations allow the model to better understand long term forecasting patterns. Another important takeaway is that the results for time series only methods are much better on SO-fore$_{2-1}$, due to the localized temporal information and shorter forecasting horizon, while the images become less important than in the previous case. 
Additionally, we tested Cross-Attention RNN in the \emph{Demand forecasting} task, i.e., predicting the full sales series without having access to any previous observations, but only to the product image. We can observe that even in this more challenging setting, results are inferior to the SO-fore variants but better than the kNN approaches. 

Our dataset and experiments provide a general overview of how problems in the fashion realm can be tackled and how the use of computer vision and multi-modal approaches is key to providing better solutions. The dataset page will contain further information regarding other, possible challenges and tasks for \dname.

\begin{table}
\footnotesize
    \centering
    \begin{tabular}{|l|c|c|}\hline
        Method & WAPE & MAE\\\hline\hline
        \textbf{\textit{Demand} CrossAttnRNN w/ image} & \textbf{89.02} & \textbf{0.99}\\\hline\hline
        Naive & 118.176 & 1.31 \\\hline
        SES & 111.265 &  1.23\\\hline
        kNN & 91.13 &1.01\\\hline
        kNN + image & 97.97 &1.06\\\hline
        CrossAttnRNN & 87.39 & 0.97 \\\hline
        \textbf{CrossAttnRNN w/ image} & \textbf{85.65} & \textbf{0.94}\\\hline
        
    \end{tabular}
    \caption{Results for SO-fore$_{2-10}$, showing the Weighted Average Percentage Error (WAPE) and Mean Absolute Error (MAE) \cite{skenderi2021well} for the different baselines. The lower the better for both metrics. In the first row we also report results for the demand forecasting of new products without past sales, demonstrating how much the knowledge of the initial sale dynamics improves forecasting.}
    \label{tab:st-res39}
    \vspace{-0.3cm}
\end{table}

\begin{table}
\footnotesize
    \centering
    \begin{tabular}{|l|c|c|}\hline
         Method & WAPE & MAE\\\hline\hline
        Naive & 101.922 & 1.13\\\hline 
        SES & 97.85 & 1.08\\\hline
        kNN & 91.11 & 1.01 \\\hline
        kNN + image & 92.97 & 1.02\\\hline
        CrossAttnRNN & 87.86 & 0.97\\\hline %& 35.21 & 0.39 % & 35.23 & 0.39
        \textbf{CrossAttnRNN w/ images} & \textbf{86.71} & \textbf{0.96}\\\hline %\textbf{34.73} & \textbf{0.38} & 32.59 & 0.36
    \end{tabular}
    \caption{Results for SO-fore$_{2-1}$, showing the Weighted Average Percentage Error (WAPE) and Mean Absolute Error (MAE) \cite{skenderi2021well} for the different baselines. The lower the better for both metrics.}
    \label{tab:st-res31}
    \vspace{-0.5cm}
\end{table}

\vspace{-0.55cm}
\paragraph{Acknowledgements}This work was partially supported by the Italian MIUR through PRIN 2017 - Project Grant 20172BH297: "I-MALL - improving the customer experience in stores by intelligent computer vision" and the MIUR project "Dipartimenti di Eccellenza 2018-2022". 

\twocolumn[{%
\renewcommand\twocolumn[1][]{#1}%
\maketitle
\begin{center}
    \centering
    \captionsetup{type=figure}
   \includegraphics[width=\linewidth]{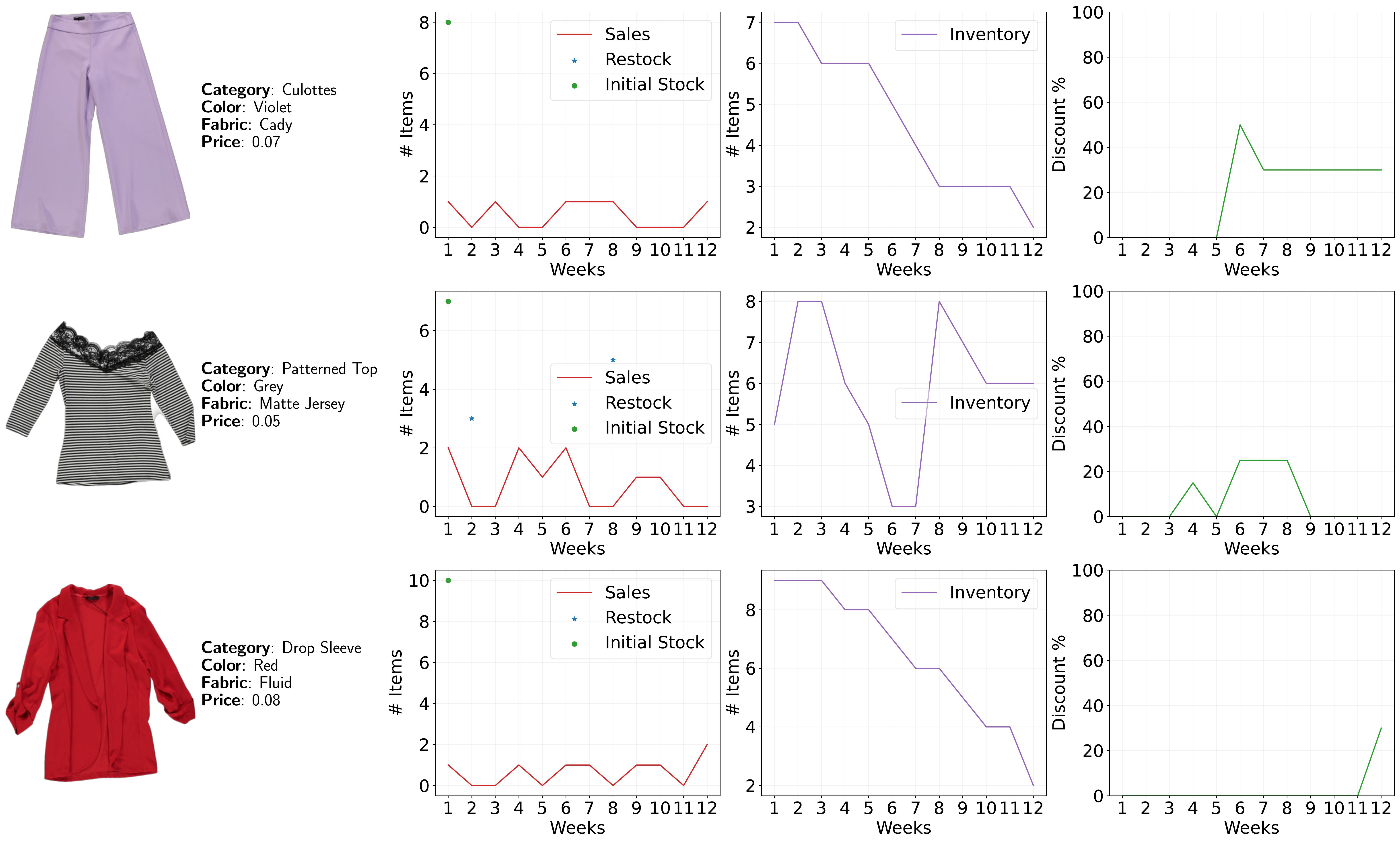}
    \captionof{figure}{Examples of different products that have been sold in the shop SHOP11 during the AW19 season. The figure reports (from left to right): the product's image; textual tags; sales time series; restocking information; inventory time series; and discount time series.}\label{fig:AW19}
\end{center}%
}]

\section{Appendix} \label{sec:outline}
The main paper discusses two contributions: 1) the novel \dname{} dataset and 2) the advantage of utilising a multi-modal approach for short-observation new product forecasting (SO-fore). This additional material adds to each one of these main topics, as follows:

\myparagraph{The \dname{} dataset} 
The following topics will be faced in Sec.~\ref{sec:dataset}:
\begin{itemize}
    \item How \dname{} compares with respect to the current datasets of forecasting for fast-fashion, exploring the literature of time-series forecasting and computer vision in this area (Sec.\ref{sec:SoA}); 
    \item A showcase of the data  contained in \dname{}, including examples of products, the associated \emph{time series data}, \emph{image data} and the \emph{text data}. An excerpt of these data is reported in Fig.~\ref{fig:AW19}. Subsequently, we will show some examples of \emph{exogenous data} i.e., the Google Trends associated to a given product, and the weather reports associated to a given shop; finally, \emph{customer purchase data} will be presented (Sec.~\ref{sec:showcase});
    \item A list of possible challenges that can be studied on \dname{} will be presented, and specifically: more tasks related to \emph{new product short observation forecasting}; the \emph{new product demand forecasting}; the \emph{product recommentation} (Sec.~\ref{sec:challenges}).
\end{itemize}

\myparagraph{The SO-fore problem} 
In Sec.\ref{sec:fore}, the Cross-Attention RNN will be detailed, with a graphical representation that illustrates its components (Sec.~\ref{sec:RNN}).

\subsection{The \dname{} dataset}\label{sec:dataset}
\subsubsection{Datasets for forecasting in fast fashion} \label{sec:SoA} The \dname{}
 dataset can be related to two scientific fields: 1) the one of forecasting for (fast) fashion~\cite{nenni2013demand,taplin2014global,choi2014fast}, with particular emphasis on those works that exploit deep learning techniques ~\cite{loureiro2018exploring,ekambaram2020attention}; 2) the recent field at the intersection between computer vision and fashion~\cite{cheng2021fashion}, with special emphasis on the task of \emph{popularity prediction} and \emph{fashion forecasting}. In both cases, \dname{} innovates for specific reasons which will be detailed in the following, in two separate sections. In both the cases, \dname{} has an unprecedented richness of data which other datasets do not possess, such as customer purchase information, and the different exogenous data. 
 
 \paragraph{\textbf{Forecasting for (fast) fashion}}
 The most used forecasting techniques for fast fashion incorporate classical ARIMA, SARIMA, exponential smoothing~\cite{brown2004smoothing}, regression~\cite{papalexopoulos1990regression}, Box \& Jenkins~\cite{box2015time} and Holt Winters~\cite{winters1960forecasting}, as reported in the recent review of~\cite{loureiro2018exploring}; machine learning approaches (decision trees, random forests, SVMs, neural networks) are at their infancy on this topic and most importantly they are not considering multi-modal data, but only time-series. This has naturally created an abundance of datasets for time-series analysis for sale forecasting~\cite{pavlyshenko2016linear} and demand forecasting~\cite{pavlyshenko2018using}, and the absence of datasets with images included, which we are filling with \dname{}. As a notable exception, the work of~\cite{ekambaram2020attention} proposes a set of techniques for demand forecasting, in which images are taken into account by an Attention-based RNN framework, which we also utilise for SO-fore as explained in Sec.~\ref{sec:RNN}. Unfortunately, the dataset on which they perform their experiments is not publicly available, while \dname{} will be made publicly  available.

 \paragraph{\textbf{Intersection between computer vision and fashion}} The peculiarity of this field is the exploration of multi-modal data (images, text, time-series) for prediction tasks. In general, computer vision approaches have been considered for the task of \emph{popularity prediction} or \emph{fashion forecasting}~\cite{cheng2021fashion} In both the cases, the ground truth signal is built on top the public ratings obtained on online platforms such as Chictopia.com~\cite{yamaguchi2014chic,simo2015neuroaesthetics}, Lookbook.nu~\cite{lo2019dressing}or Amazon~\cite{al2017fashion}, which consider outfits~\cite{simo2015neuroaesthetics,yamaguchi2014chic,simo2015neuroaesthetics,lo2019dressing}, or several outfits exhibiting the same style~\cite{al2017fashion}: an outfit or a style is popular if it receives a high rating in terms of number of ``likes'' or ``stars''. In the case of \dname{}, we can assume that a product is more popular than others of the same category, if in the same season, it has sold more. In this setup, our dataset allows to be more fine-grained, since one can predict the popularity, in terms of sales, of a single product. Also, \dname{} represents the very first dataset which permits multi-modal analysis on the data of a real fast fashion company, meaning that approaches which succeed on this benchmark can be directly applicable on the fast fashion market.      

\subsubsection{\dname{}: a showcase}\label{sec:showcase}

\myparagraph{Example of products}
In Sec.2 of the main paper we have given some statistics about the products which are in the \dname{} dataset. Here we will give some qualitative examples of their \emph{image data}, \emph{text attributes} and associated time series: \emph{product sales}, \emph{inventory position}, \emph{Restock flag}, and \emph{discount}, the latter omitted in the main paper due to the lack of space. Formally, following the notation of the main paper, given a product $i$ at a retail store $r$, we refer to its \emph{discount} signal as $D(i,r,t)$ where $t$ refers to the $t$-th week of market delivery, with $i=1,...,N$, $r=1,...,L$ and $t=1,...,K$.  The discount signal is expressed as a percentage, describing how much a particular item is discounted; for example, $D(i,r,t)=20$ indicates that the initial price defined for a product $i$ in the retail store $r$ was discounted by a 20\% at time $t$.  Fig.~\ref{fig:AW19} showcases all of these data for three products of season AW19. Other figures can be found at the end of this additional material, reporting products of other seasons (Fig.~\ref{fig:exampless17}, Fig.~\ref{fig:exampleAW17} and Fig.~\ref{fig:exampless18}).

\begin{figure}
    \centering
    \includegraphics[width=\linewidth]{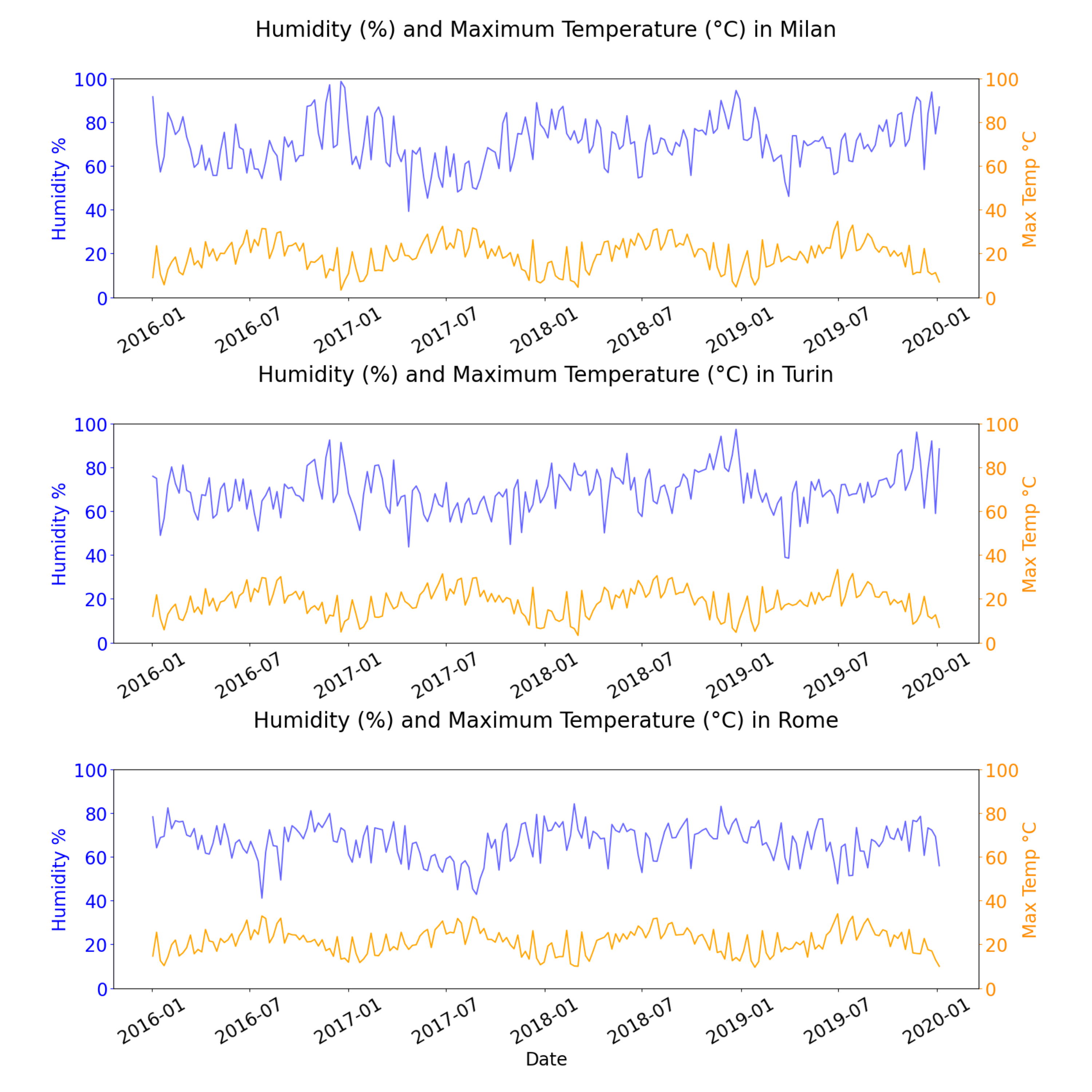}
    \caption{Excerpt of exogenous weather data (humidity and maximum temperature) for three major Italian municipalities: Milan, Turin and Rome. Humidity in these areas tends to be negatively correlated with maximum temperature. Rome tends to be much hotter than Milan and Turin throughout the year. An interesting observation which can prove beneficial to forecasting is the seasonal nature of weather phenomena, which can induce information as to which clothing products may sell more in a particular period.}
    \label{fig:weather}
\end{figure}

\begin{figure*}
    \centering
    \includegraphics[width=\linewidth]{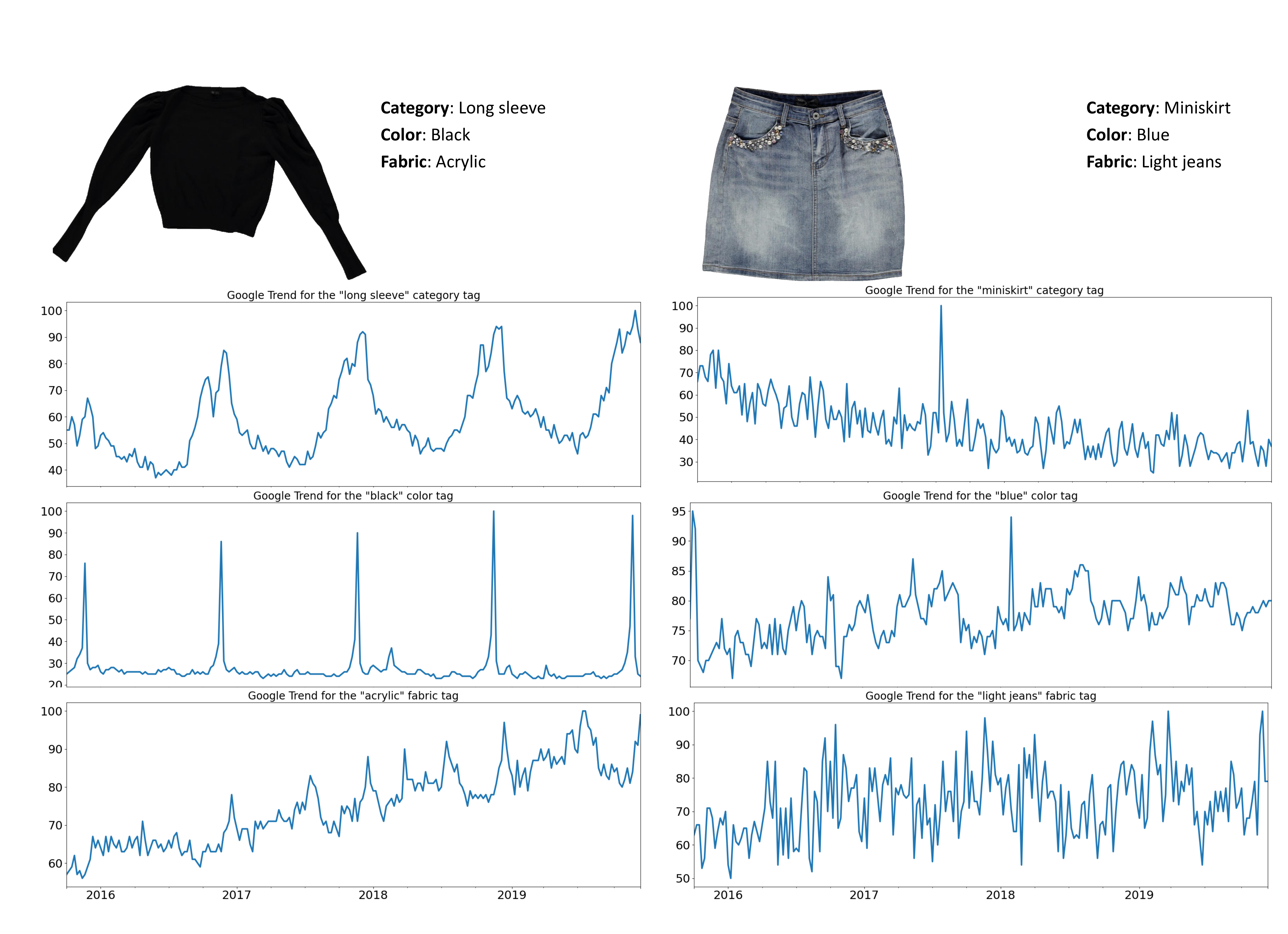}
    \caption{Example of the exogenous Google Trends data available in \dname for two completely different products. The signals can have regular trend and seasonality (left) or be stationary and seem noisy (right). This can prove helpful for forecasting models, because it allows to understand both global and cyclical popularity and therefore anticipate sales.}
    \label{fig:gtrends}
\end{figure*}

\myparagraph{Exogenous data}
Exogenous data is often neglected as a resource within datasets, especially in forecasting. This is due to their nature, since they are, by definition, coming from an external phenomenon that is not directly related with the data being analysed. Nevertheless, adding exogenous variables such as weather data \cite{beheshti2015survey,sull2008fast} or popularity data \cite{silva2019googling,skenderi2021well} to forecasting models has proven extremely beneficial in terms of forecasting performance. For this reason, we provide in \dname multivariate exogenous data both for the weather and popularity, in the form of detailed weather reports (Fig. \ref{fig:weather}) and Google Trends (Fig. \ref{fig:gtrends}). An more profound explanation for both examples is provided in the respective captions.

\begin{figure*}
    \centering
    \includegraphics[width=\linewidth]{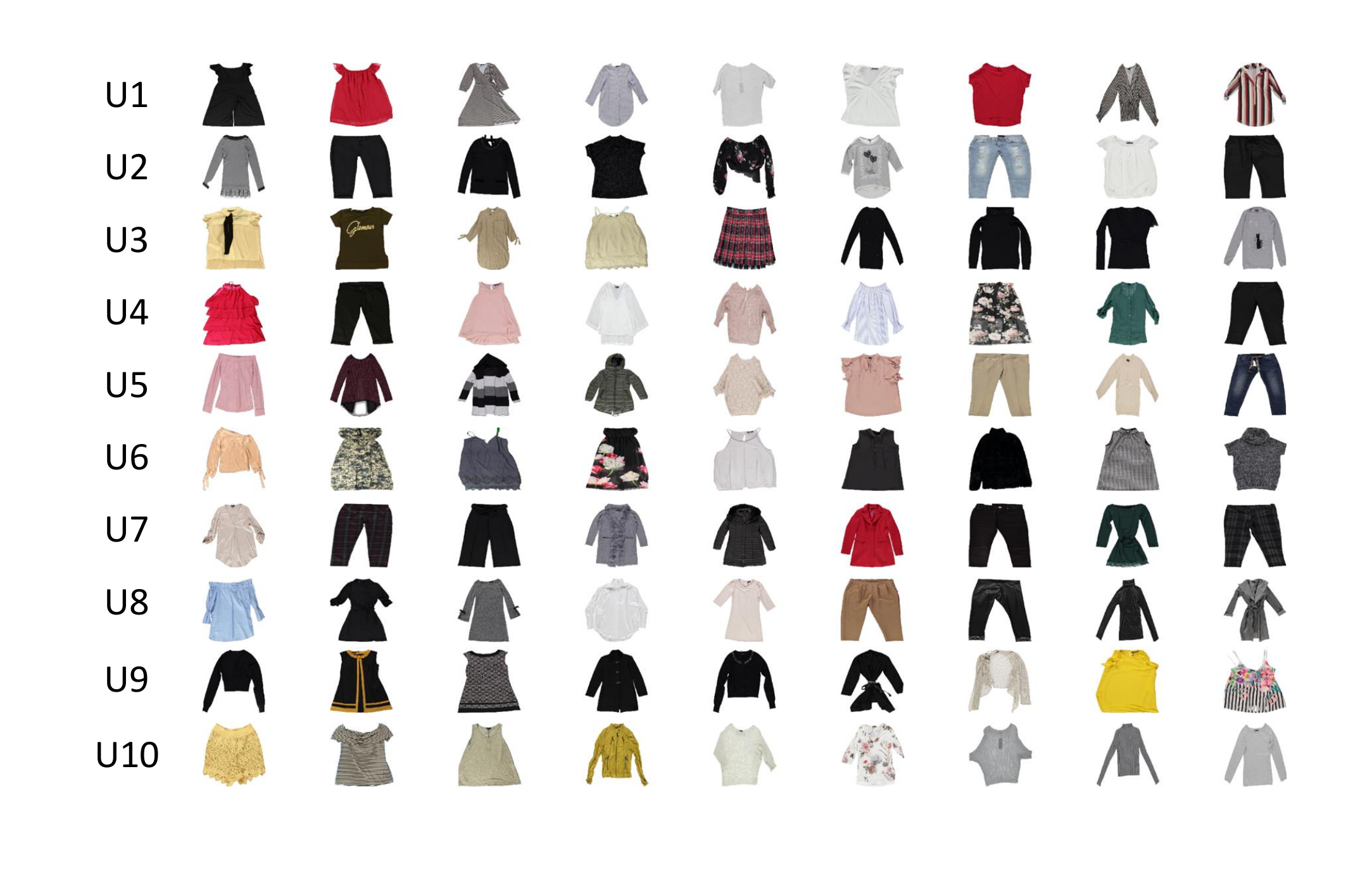}
    \caption{A random sampling of users/ purchases. Personal styles do emerge: users 1 and 10 have no trousers in their logs, user 6 has bought almost short sleeves and no trousers, while user 7 seems to prefer long sleeves and several trousers; user 10 has a marked preference for light yellow-grayish colors.}
    \label{fig:example_cust1}
\end{figure*}

\myparagraph{Customer purchase data}
Quoting Sec.2 and Fig.6 of the main paper, among the 667086 total registered users of Nuna Lie, 6k users have bought continuously a total of 25 products over 4 seasons. In Fig.~\ref{fig:example_cust1}
we have a random excerpt of 10 of these users, with a random subset of 9 purchased items each (no cherry picking); in many cases, personal styles do emerge, showing systematic preferences on diverse attributes, as written in the caption of the figure.

\subsubsection{Challenges on \dname{}}\label{sec:challenges}
In this paper, we explored the problem of short-observation forecasting on \dname{}, with the precise focus of showing the benefit of the image data on this task. Obviously, we are far from saturating the performance, encouraging further improvements. These could be provided by diverse techniques (exploiting LSTM or transformer-based architectures), or including additional exogenous data, available in \dname{}. Google Trends and weather reports, in fact, are signals which have been shown elsewhere to be predictive~\cite{beheshti2015survey,sull2008fast,silva2019googling,skenderi2021well}, so this should be a natural next step. Other challenges which can be experimented on \dname{} are listed in the following.
\begin{itemize}
    \item \textbf{Demand forecasting.} Forecasting demand is a crucial issue for driving efficient operations management plans~\cite{nenni2013demand,thomassey2016intelligent,feizabadi2022machine}.This is especially the case in the fast fashion industry,where demand uncertainty, lack of historical data, variable ultra-fast life-cycle of a product and seasonal trends usually coexist~\cite{maass2014improving,kalla2020improved}. In rough terms demand forecasting outputs the amount of goods to buy from the suppliers. This amount is then distributed among the different retailers, with the aim of avoiding zero-stocks or excessive unsold inventory. In this paper we show a glimpse of demand forecasting on \dname{}, at the level of single shop (i.e. predicting how much a single shop will need during the next season), adopting the recent RNN-based approach of~\cite{ekambaram2020attention} on time series and time series + image. In~\cite{skenderi2021well} we report some results on the aggregated signal in the old Visuelle dataset; it is worth remembering that in that case the signal about the single shops was missing, less products were available, and the only important result was to show how Google Trends data are beneficial.  In this case, a deep analysis on demand forecasting on \dname{} needs to be carried out, including discounts and exogenous signals like weather reports (Fig. \ref{fig:weather}) and Google Trends (Fig. \ref{fig:gtrends}). 
    
    \item \textbf{Product recommendation.} An important feature of \dname{} is the presence of customer purchase data; 667086 customers have bought along 8 seasons a total amount of 3253876 items, which cover a consistent percentage (84\%) of the total purchases collected within the dataset. A graphical representation of these data is reported in Fig.~\ref{fig:example_cust1}, where it is visible that some users have marked preferences.
    
    Product recommendation on these data would consist in defining a particular time index $t_{\text{rec}}$, when the historical data of all the past purchases (older than $t_{\text{rec}}$) of all the customers will be taken into account. Therefore, two types of inferences will be possible: 1) to suggest which product (or category, or attribute) $z_{k}$ a specific customer $u_{i}$ could be interested in; a positive match will be in the case of an effective purchase of $z_{k}$ (or some item which is in the category $z_{k}$ or that expresses the attribute $z_{k}$) by $u_{i}$ after time $t_{\text{rec}}$; 2) same as before, but including a specific time interval $T_{\text{buy}}$ within which the customer will buy. In practice,  a positive match will be in the case of an effective purchase of $z_{k}$ (or some item which is in the category $z_{k}$ or that expresses the attribute $z_{k}$) by $u_{i}$ \emph{in the time interval} $]t_{\text{rec}},t_{\text{rec}}+T_{\text{buy}}]$.
In general, product recommendation can be carried out by standard collaborative-filtering based techniques, but also considering recommendation as an instance of forecasting~\cite{hu2015web,fayyaz2020recommendation} and viceversa: this interplay could be certaintly explored with the \dname{} dataset.  
\end{itemize}

\subsection{The SO-fore problem}\label{sec:fore}
\subsubsection{Cross-Attention RNN} \label{sec:RNN}
The Cross-Attention RNN \cite{ekambaram2020attention}, can be described as an autoregressive, sequence-to-sequence neural network that tries to understand the different, non-linear relationships in the various data modalities and then perform predictions by understanding which part of the data is most important for the forecasting task. The attention modules constitute a large part of the model and are exactly as in \cite{Bahdanau2015}, where at each decoding step we try to attend to the encoder features based on the current decoder hidden state.
\begin{figure}[]
    \centering
    \includegraphics[width=0.7\linewidth]{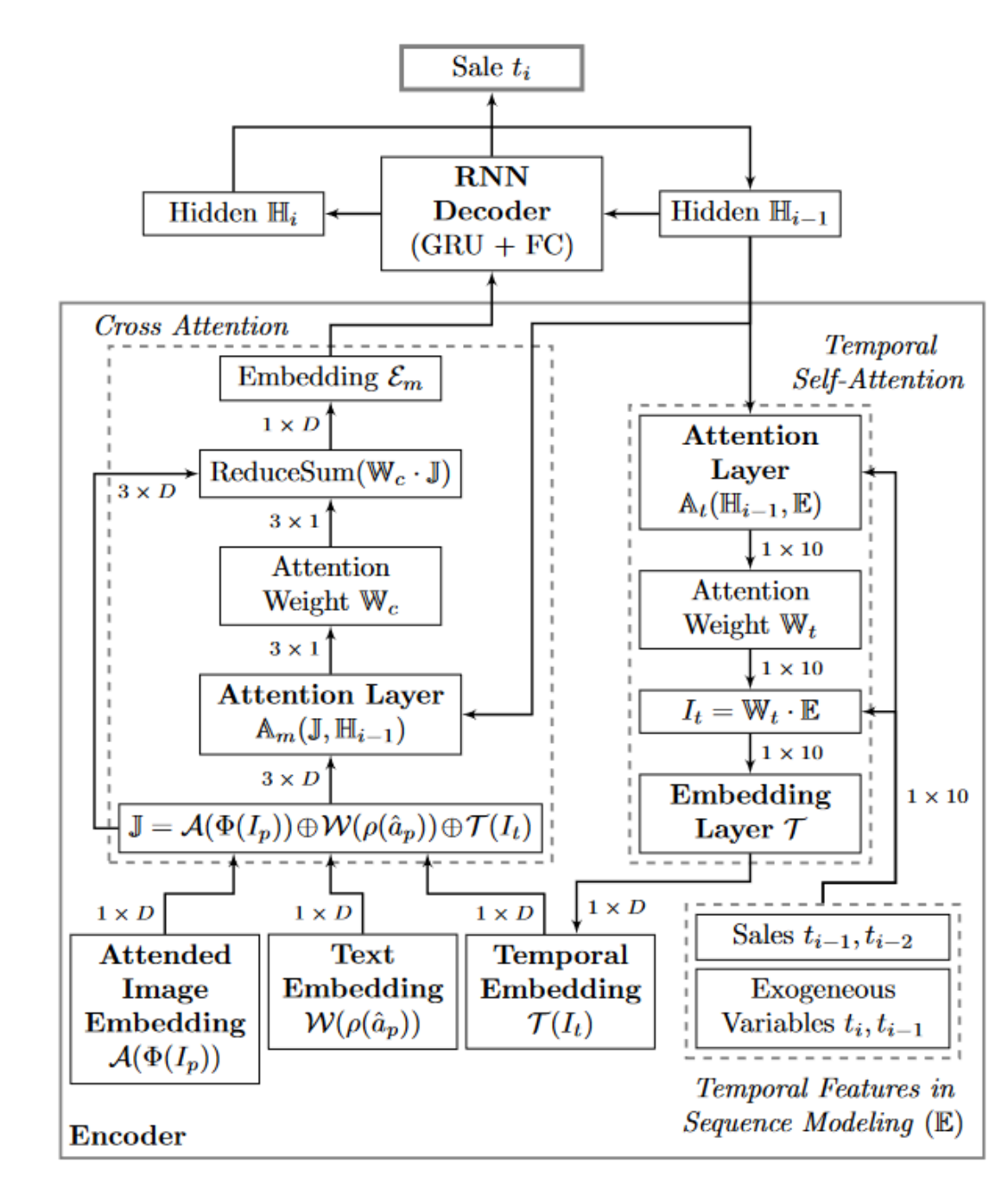}
    \caption{A visual description of the Cross-Attention RNN model, the neural network architecture used in our SO-fore experiments. The model diagram is taken from\cite{ekambaram2020attention}.}
    \label{fig:rnnmodel}
\end{figure}

The encoder starts by embedding each input modality into a common feature space $R^D$. The input observation sales (time series) are first passed through an additional self-attention layer \cite{vaswani_nips2017}, differently from the original work in \cite{ekambaram2020attention} and then projected through a fully connected layer. This helps filter out initial noise from past sales observations. The image and textual tags are embedded processed using a ResNet-101 \cite{He_2016_CVPR} and learnable embedding layers respectively. The temporal features extracted from the product's release date are also embedded using learnable embedding layers.

Cross-attention RNN works, by default, in an autoregressive manner, therefore at each decoding step three different additive attention modules are applied. These modules allow the decoder hidden state to attend to the time series embedding, the image embedding and most importantly to the concatenated, multi-modal embedding. A residual learning approach \cite{He_2016_CVPR} is applied to allow the network to scale better with the number of hidden layers and also learn to ignore null contributions from the attention mechanism. After each decoding step, the GRU hidden state is updated based on the last processed input. For extensive details on this model, we refer to \cite{ekambaram2020attention}.

The model is trained with a batch size of 128 and MSE (Mean Squared Error) loss function, using the Adafactor optimizer, on two NVIDIA RTX TITAN GPUs. During training, we apply dropout after each embedding module and also apply teacher forcing at random with a probability $p_{tf}=0.5$.

\begin{figure*}[b!]
    \centering
    \includegraphics[width=0.95\linewidth]{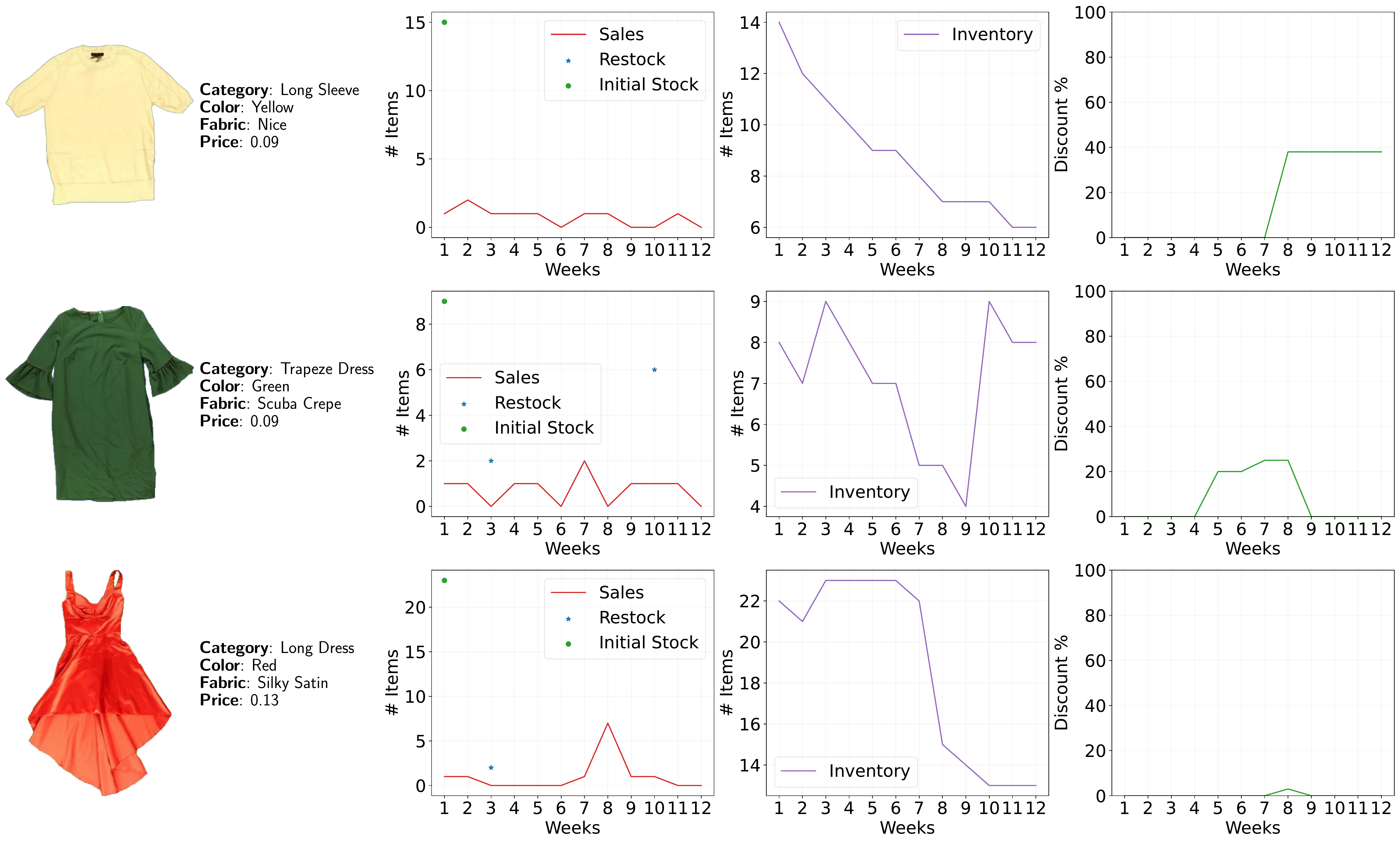}
    \caption{Examples of different products that have been sold in the shop SHOP3 during the SS17 season. The figure reports (from left to right): the product's image; textual tags; sales time series; restocking information; inventory time series; and discount time series.}
    \label{fig:exampless17}
\end{figure*}

\begin{figure*}
    \centering
    \includegraphics[width=0.95\linewidth]{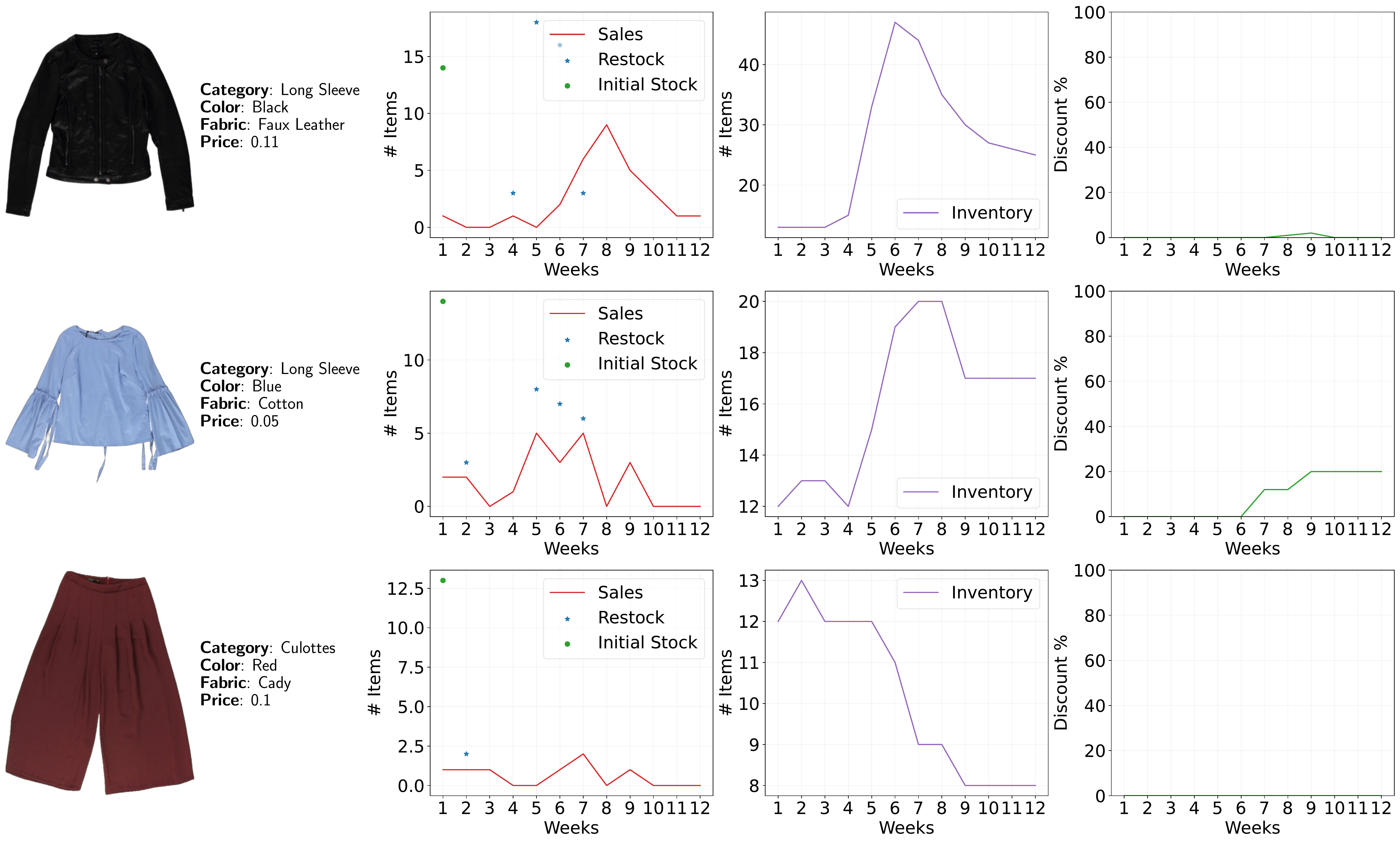}
    \caption{Examples of different products that have been sold in the shop SHOP28 during the AW17 season. The figure reports (from left to right): the product's image; textual tags; sales time series; restocking information; inventory time series; and discount time series.}
    \label{fig:exampleAW17}
\end{figure*}

\begin{figure*}
    \centering
    \includegraphics[width=0.95\linewidth]{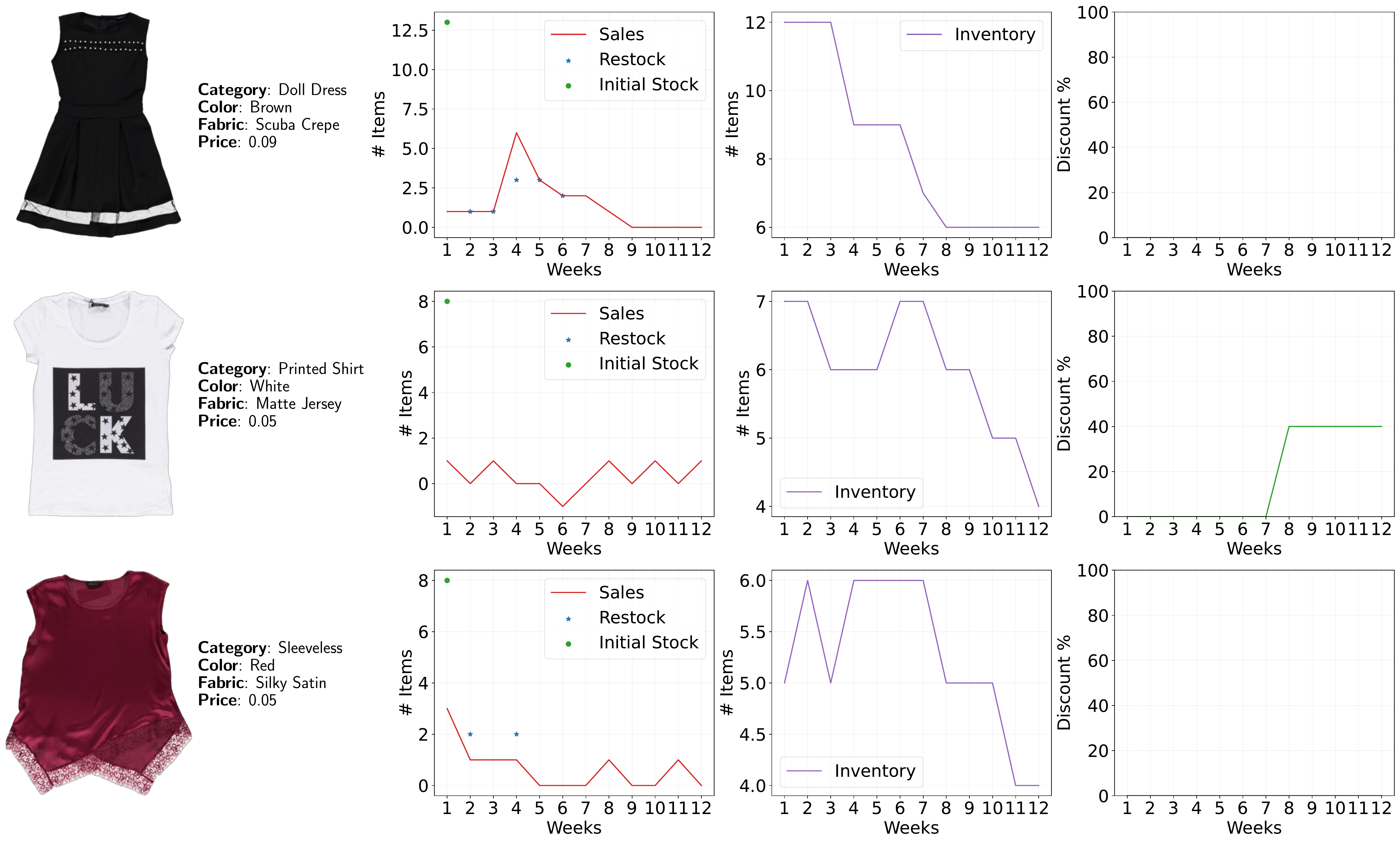}
    \caption{Examples of different products that have been sold in the shop SHOP51 during the SS18 season. The figure reports (from left to right): the product's image; textual tags; sales time series; restocking information; inventory time series; and discount time series.}
    \label{fig:exampless18}
\end{figure*}

%%%%%%%%% REFERENCES
\clearpage
{\small
\bibliographystyle{ieee_fullname}
\bibliography{egbib}
}

\end{document}